\documentclass[conference]{IEEEtran}
\usepackage{graphicx,times,psfig,amsmath,amscd,amssymb,mathrsfs} 

\newtheorem{theorem}{Theorem}

\newtheorem{example}[theorem]{Example}
\newtheorem{remark}[theorem]{Remark}

\newcommand{\abs}[1]{\lvert#1\rvert}

\def\s{{\mathbb S}}

\def\I {{\mathbb I}}

\def\R{{\mathbf R}}

\def\e{\varepsilon}
\def\me{\mathrm{me}_1}

\IEEEoverridecommandlockouts	
				
\begin{document}

\title{\ \\ \LARGE\bf Intrinsic dimension of a dataset: what properties does one expect?
\thanks{Vladimir Pestov is with the Department of Mathematics and Statistics, University of Ottawa, 585 King Edward Avenue, Ottawa, Ontario, K1N 6N5 Canada (phone: 613-562-5800 ext. 3523, fax: 613-562-5776, email: vpest283@uottawa.ca).}}

\author{Vladimir Pestov}
\maketitle

\begin{abstract}
We propose an axiomatic approach to the concept of an intrinsic dimension of a dataset, based on a viewpoint of geometry of high-dimensional structures. Our first axiom postulates that high values of dimension be indicative of the presence of the curse of dimensionality (in a certain precise mathematical sense). The second axiom requires the dimension to depend smoothly on a distance between datasets (so that the dimension of a dataset and that of an approximating principal manifold would be close to each other). The third axiom is a normalization condition: the dimension of the Euclidean $n$-sphere $\s^n$ is $\Theta(n)$. We give an example of a dimension function satisfying our axioms, even though it is in general computationally unfeasible, and discuss a computationally cheap function satisfying most but not all of our axioms (the ``intrinsic dimensionality'' of Ch\'avez et al.) 
\end{abstract}

\section{Introduction}

A search for the ``right'' concept of intrinsic dimension of a dataset is not yet over, and most probably one will have to settle for a spectrum of various dimensions, each serving a particular purpose, complementing each other. (Cf. \cite{cv,CNBYM,ch,Kegl,pa,tmgm,ttf}, and references therein.) At the same time, it is quite clear that the word ``dimension'' has a rather specific meaning in this context. High values of dimension are invariably associated with the curse of dimensionality, while the low values are expected to contain useful information, for instance, about a non-linear manifold approximating the dataset. Is it too much to expect of a dimension function?

Here we are trying to address the problem of existence of dimension functions making sense for all datasets and satisfying the above two requirements, within the contraints of a certain mathematical model.
Datasets are modelled by spaces $(X,d,\mu)$ equipped with a distance $d$ and a probability distribution $\mu$, while features of datasets correspond to $1$-Lipschitz (non-expanding) functions $f$ on $X$. The curse of dimensionality describes a situation where the features are sharply concentrated around their means. In geometric terms, one speaks here of the phenomenon of concentration of measure on high-dimensional structures \cite{Pe00}. This phenomenon admits well-understood quantitative measures \cite{MS,Gr,L}, which enable us to express in precise mathematical terms the following condition on an instrinsic dimension function: high values of dimension are indicative of the presence of the curse of dimensionality. 

Geometry of high dimensions (asymptotic geometric analysis) has in store a concept of a distance between spaces with metric and measure, $X$ and $Y$, which, in our view, could --- in one form or other --- eventually become very useful in principal manifold theory. We describe this notion, due to Gromov \cite{Gr}, and state the second axiom: if the Gromov distance between two spaces is small, their intrinsic dimensions should be close to each other.

The third axiom serves a normalization purpose by stating that the intrinsic dimension of the Euclidean sphere $\s^n$ should be on the order of $n$. 

Paradoxically, any dimension function of the suggested kind always assigns to a singleton the value $+\infty$, however this does not lead to any problems or contradictions.

We give an example of a dimension function satisfying the axioms, and compute its values for the spheres $\s^n$. In general, however, this function is computationally unfeasible. We discuss in this connection the ``intrinsic dimensionality'' by Ch\'avez {\em et al.}, easy to compute and already having uses in data engineering \cite{CNBYM}, which satisfies some, but not all, of our axioms. 

\section{Preliminaries}

\subsection{Metric spaces with measure as models for datasets}
A geometric model for a dataset \cite{Pe99,Pe00} is a {\em metric space with measure} \cite{MS,Gr}, that is, a triple $(X,d,\mu)$, where $X$ is a set equipped with a metric $d$ and a probability measure distribution $\mu$.
Sometimes $\mu$ is thought of as an underlying distribution for the actual set of data, else one can associate to $X$ the normalized counting measure $\mu(A)=\sharp(A)/\sharp(X)$.

In some situations, especially in sequence-based biology, a metric $d$ has to be replaced with a more general similarity measure between datapoints, such as a quasimetric \cite{PeSt06}.

\subsection{$1$-Lipschitz functions as models for features}
{\em Features} of datasets correspond in the above setting to functions $f$ on $X$ taking values in the real numbers, the Euclidean space, or another target space (such as e.g. a discrete set). The features are assumed to depend smoothly on the distance between datapoints. After a suitable normalization, one can usually assume such a function, $f$, to be {\em $1$-Lipschitz:} for all $x,y\in X$, one has 
\[\abs{f(x)-f(y)}\leq d(x,y).\]
The features are in a sense the ``observable quantities'' of a dataset. 

\subsection{\label{ss:C}Observable diameter and concentration phenomenon}

The curse of dimensionality is a name given to the situation where all or some of the important features of a dataset sharply concentrate near their median (or mean) values and thus become non-discriminating. In such cases, $X$ is perceived as intrinsically high-dimensional. This set of circumstances covers a whole range of well-known high-dimensional phenomena such as for instance sparseness of points (the distance to the nearest neighbour is comparable to the average distance between two points \cite{BGRS}), etc. It has been argued in \cite{Pe00} that a mathematical counterpart of the curse of dimensionality is the well-known {\em concentration phenomenon} \cite{M00,L}, which can be expressed, for instance, using Gromov's concept of the {\em observable diameter} \cite{Gr}.

Let $(X,d,\mu)$ be a metric space with measure, and let $\kappa>0$ be a small fixed threshold value. 
The {\em observable diameter} of $X$ is the smallest real number, $D={\mathrm{ObsDiam}}_\kappa(X)$, with the following property: for every two points $x,y$, randomly drawn from $X$ with regard to the measure $\mu$, and for any given $1$-Lipschitz function $f\colon X\to\R$ (a feature), the probability of the event that values of $f$ at $x$ and $y$ differ by more than $D$ is below the threshold value:
\[P[\abs{f(x)-f(y)}\geq D]<\kappa.\]
Informally, the observable diameter ${\mathrm{ObsDiam}}_\kappa(X)$ is the size of a dataset $X$ as perceived by us through a series of randomized measurements using arbitrary features and continuing until the probability to improve on the previous observation gets too small. The observable diameter has little (logarithmic) sensitivity to $\kappa$.

The {\em characteristic size} ${\mathrm{CharSize}}\,(X)$ of $X$ as the median value of distances between two elements of $X$. 
The concentration of measure phenomenon refers to the observation that ``natural'' families of geometric objects $(X_n)$ often satisfy
\[{\mathrm{ObsDiam}}_\kappa (X_n)\ll {\mathrm{CharSize}}\,(X_n)\mbox{ as }n\to\infty.\]
A family of spaces with metric and measure having the above property is called a {\em L\'evy family}. Here the parameter $n$ usually corresponds to dimension of an object defined in one or another sense.

\begin{figure}[htp]
\centerline{\scalebox{0.22}[0.271]{\includegraphics{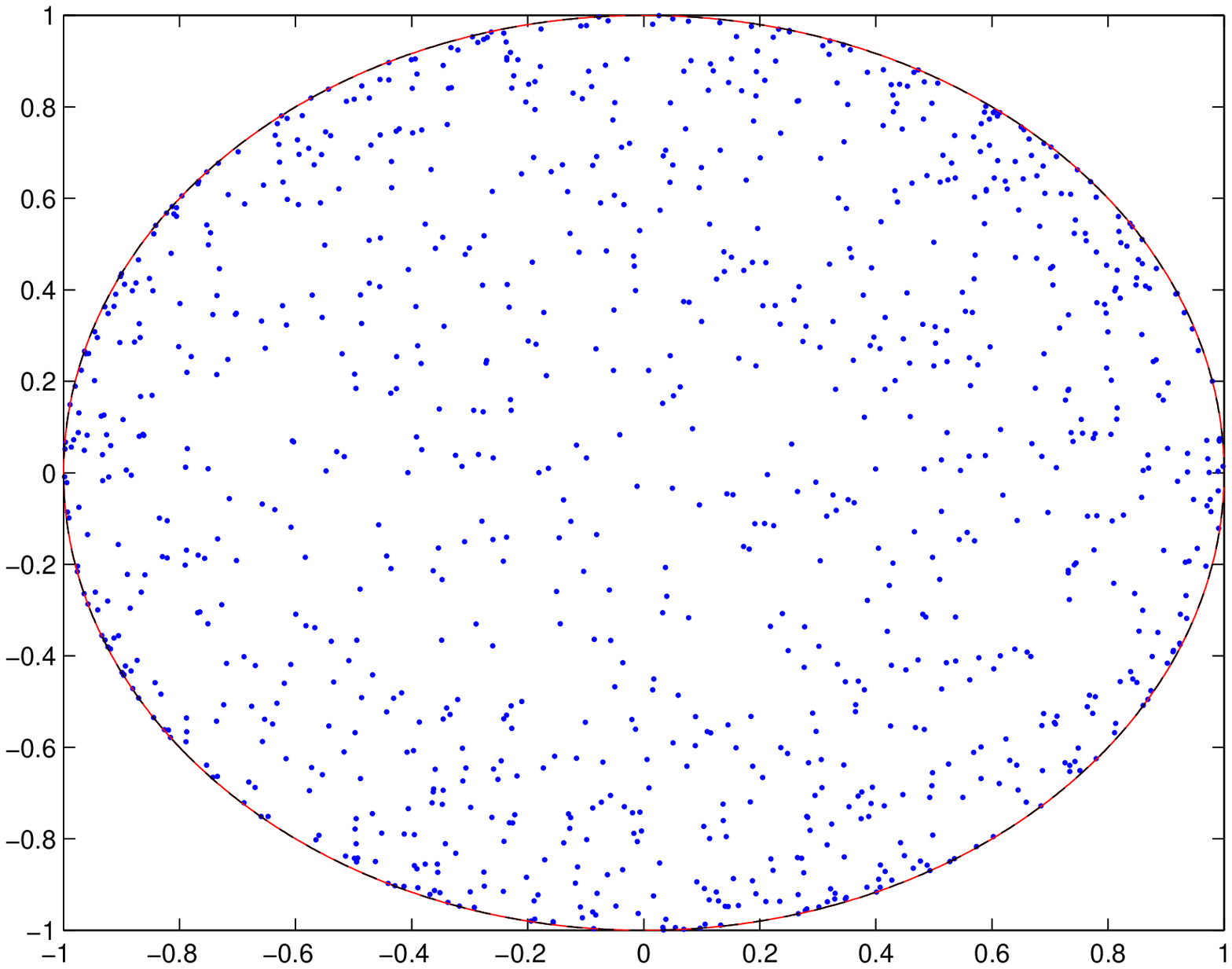}}
\hskip 1cm
\scalebox{0.22}[0.271]{\includegraphics{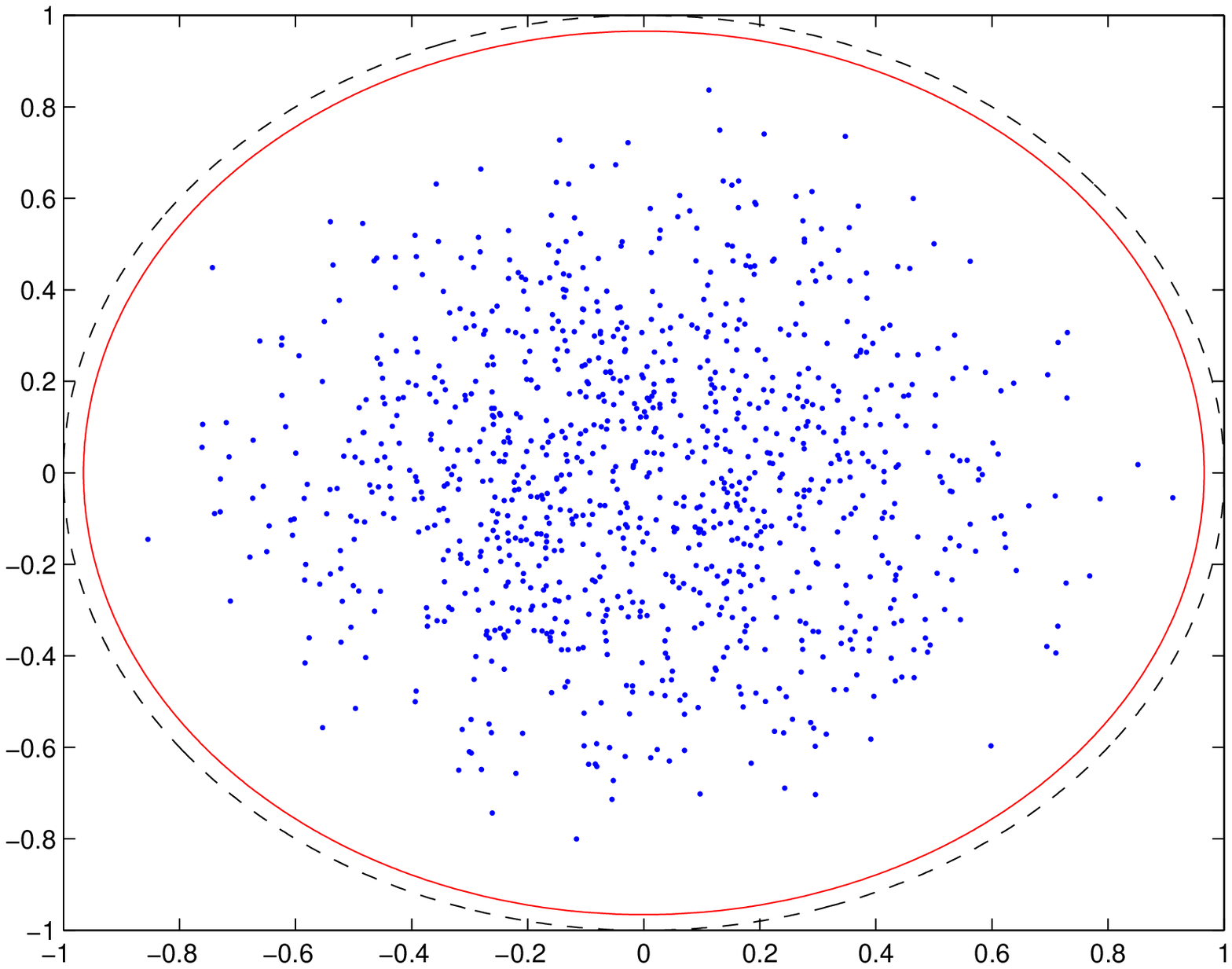}}}
\centerline{\scalebox{0.22}[0.271]{\includegraphics{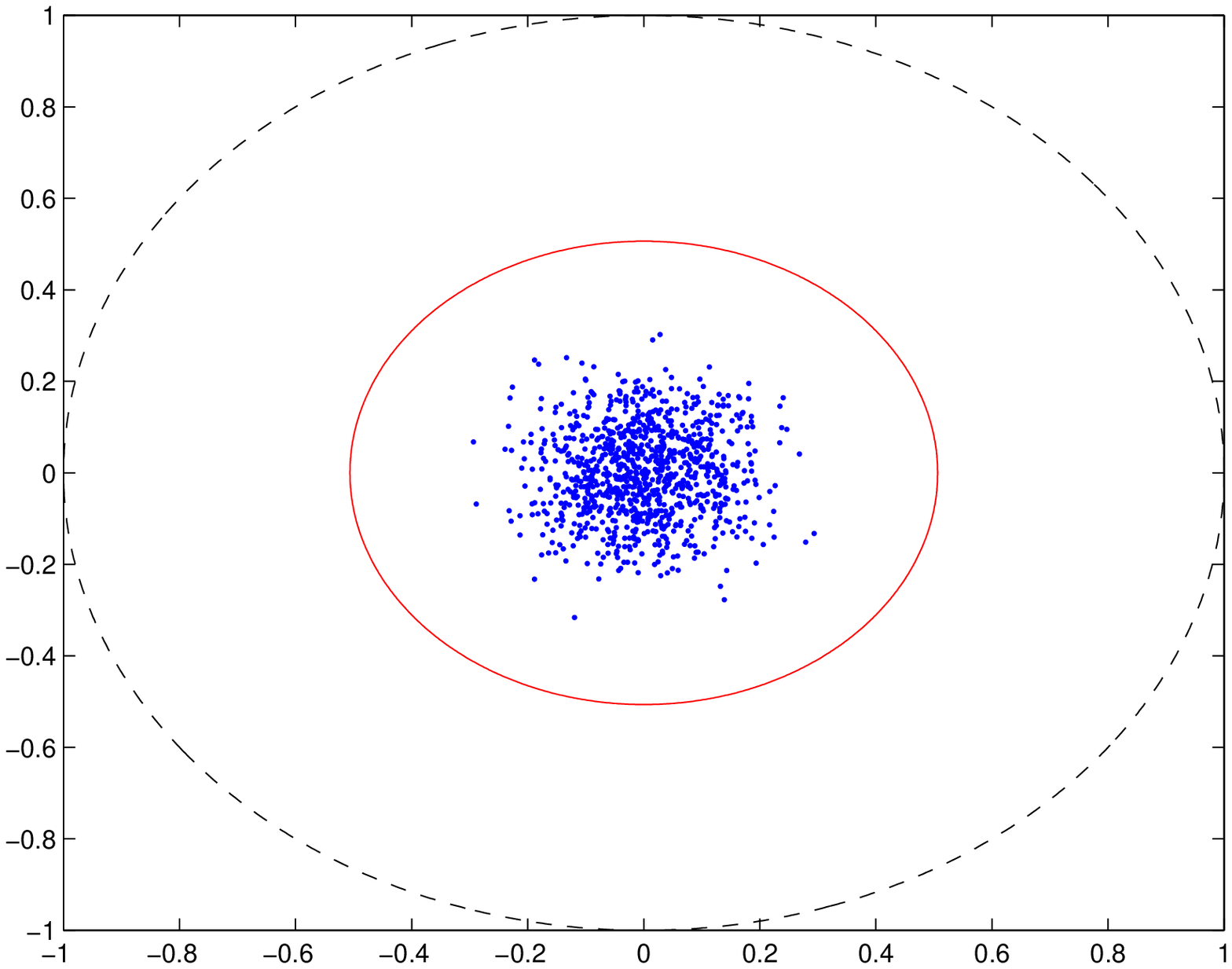}}
\hskip 1cm
\scalebox{0.22}[0.271]{\includegraphics{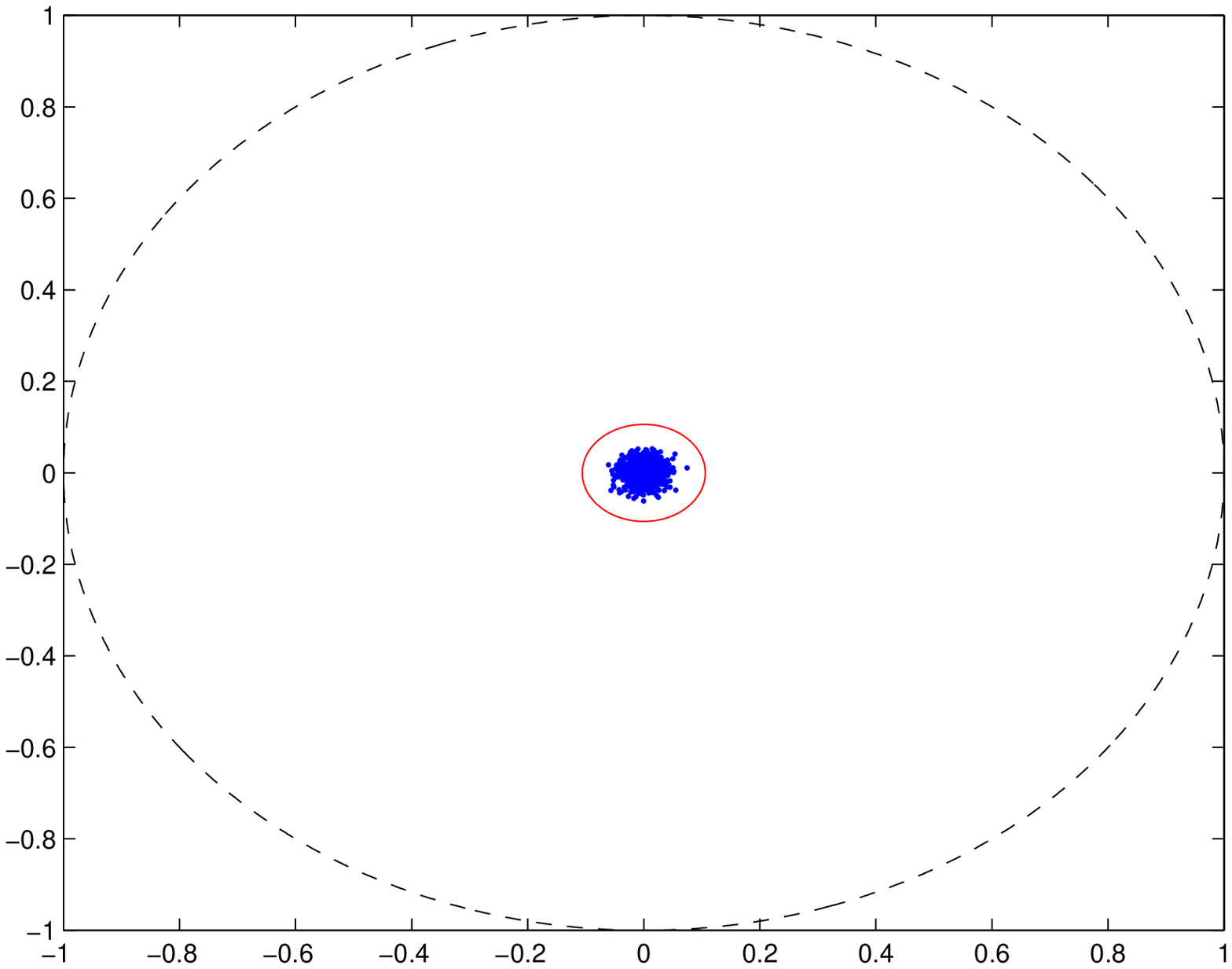}}}
\caption{Observable diameter of the sphere $\s^n$, $n=3,10,100,2500$.}
\label{fig:obs-diam}
\end{figure}

For the Euclidean spheres $\s^n$ of unit radius, equipped with the usual Euclidean distance and the (unique) rotation-invariant probability measure, one has ${\mathrm{CharSize}}(\s^n)\to \sqrt 2$, while ${\mathrm{ObsDiam}}(\s^n)=O(1/\sqrt n)$. 
Fig. \ref{fig:obs-diam} shows observable diameters (indicated by inner circles) corresponding to the threshold value $\kappa=10^{-10}$ of spheres $\s^n$ in dimensions $n=3,10,100,2500$, along with projections to the two-dimensional screen of randomly sampled 1000 points. 

Other important examples of L\'evy families \cite{MS,L,Gr}
include: (i) Hamming cubes $\{0,1\}^n$ of two-bit $n$-strings equipped with the normalized Hamming distance $d(\sigma,\tau)=\frac 1n\sharp\{i\colon \sigma_i\neq\tau_i\}$ and the counting measure; (ii) groups $SU(n)$ of special unitary $n\times n$ matrices, with the geodesic distance and Haar measure (unique invariant probability measure); (iii) any family of expander graphs (\cite{Gr}, p. 197) with the normalized counting measure on the set of vertices and the path metric.

Any dataset whose observable diameter is small relative to the characteristic size will be suffering from dimensionality curse.

\subsection{\label{ss:E}Concentration function}
A convenient way to quantify the concentration phenomenon is provided by the {\em concentration function,} $\alpha(\e)$, of a space $(X,d,\mu)$ \cite{MS,L}. Here is a definition in terms of features ($1$-Lipschitz functions). Denote by $M_f$ the median value of a function $f$, that is, a number such that
\[\mu\{x\in X\colon f(x)\geq M_f\}\geq\frac 12,~~
\mu\{x\in X\colon f(x)\leq M_f\}\geq\frac 12.\]
Now set $\alpha(0)=\frac 12$, and for every $\e>0$
\begin{equation}
\alpha(\e)=\sup\mu\left\{x\in X\colon f(x)\geq M_f+\e \right\},
\label{eq:concf}
\end{equation}
where the supremum is taken over all $1$-Lipschitz real-valued functions on $X$. Thus, the value $\alpha(\e)$ of the concentration function gives an upper bound on the probability of a large deviation of any feature from its median.
Equivalently,
\[\alpha(\e)=1-\inf \mu(A_\e),\]
where $A_\e$ denotes the $\e$-neighbourhood of $A$ in $X$ (the set of all $x$ at a distance $<\e$ to some point in $A$), and the infimum is taken over all subsets $A\subseteq X$ satisfying $\mu(A)\geq \frac 12$.

\begin{figure}[htp]
\centerline{\includegraphics[width=8.51cm]{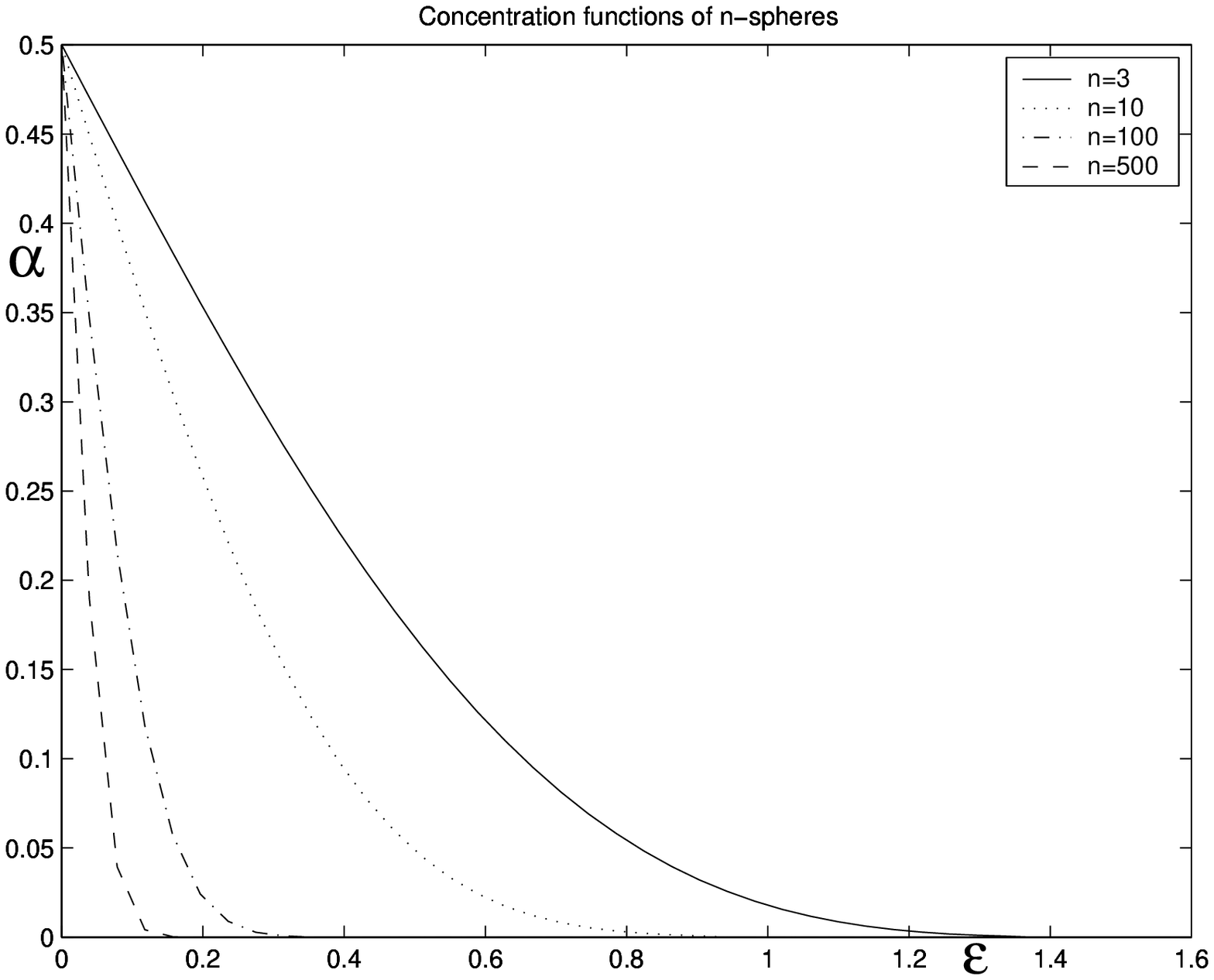}}
\caption{Concentration functions of a $n$-spheres for various $n$}
\label{fig:sph500}
\end{figure}

A family $(X_n)$ of spaces with metric and measure is a L\'evy family as defined in Subsection \ref{ss:C} if and only if the values of concentration functions $\alpha_{X_n}(\e)$ converge to zero pointwise for every $\e>0$. Concentration functions of spheres in various dimensions are shown in Fig. \ref{fig:sph500}.

\subsection{Gromov distance} 
We proceed to describe a distance between spaces with metric and measure as introduced by Gromov \cite{Gr}, p. 200.

Recall that the {\em Hausdorff distance} between two subsets $A$ and $B$ of a metric space $(X,d)$ is the smallest $\e>0$ with the property
\[A\subseteq B_\e\mbox{ and }B\subseteq A_\e.\]
(The $\e$-neighbourhood, $A_\e$, of $A$ was defined above in \ref{ss:E}.)

Let $(X,d_X,\mu_X)$ and $(Y,d_Y,\mu_Y)$ be two spaces with metric and measure. Denote by ${\mathcal{L}ip}_1(X)$ and ${\mathcal{L}ip}_1(Y)$ the spaces of $1$-Lipschitz real-valued functions (i.e., features) on $X$ and on $Y$, respectively. 
Informally, the Gromov distance between $X$ and $Y$ is the Hausdorff distance between ${\mathcal{L}ip}_1(X)$ and ${\mathcal{L}ip}_1(X)$. Of course, in order to measure it, one needs to ``pull back'' all the functions to a common third space.

This space is the function space on the unit interval $[0,1]$. It is a standard result in measure theory that every measure space $(X,\mu)$ (under mild restrictions met e.g. by every space with metric and measure) admits a {\em parametrization}, that is, a mapping $\phi\colon [0,1]\to X$ with the property: for all $A\subseteq X$, $\mu(A)$ equals the Lebesgue measure of $\phi^{-1}(A)$. For instance, if $X$ is a finite set with the normalized counting measure, then $\phi$ would be a  function taking a constant value $x\in X$ on each of $n=\sharp(X)$ intervals of equal measure. 

\begin{figure}[htp]
\centerline{\includegraphics[width=8.51cm]{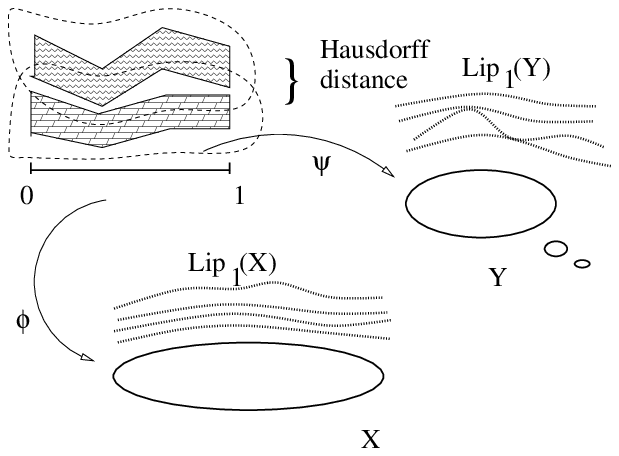}}
\caption{To the concept of Gromov's distance}
\label{fig:gromdist}
\end{figure}

Choose parametrizations $\phi$ for $X$ and $\psi$ for $Y$, and denote $\phi^\ast {\mathcal{L}ip}_1(X)$ the set of all functions of the form $f\circ \phi$, $f\in {\mathcal{L}ip}_1(X)$, and similarly $\psi^\ast {\mathcal{L}ip}_1(Y)$. Both $\phi^\ast {\mathcal{L}ip}_1(X)$ and $\psi^\ast {\mathcal{L}ip}_1(Y)$ are subspaces of the space $L^1(0,1)$ of integrable functions on the unit interval. Equip the latter space with the following metric, determining the {\em convergence in measure:}
\[\me(f,g)=\inf\left\{\e>0\colon \mu\{x\colon \abs{f(x)-g(x)}>\e\}<\e\right\}.\]
Now the Gromov distance $d_{conc}(X,Y)$ is the infimum of Hausdorff distances between the subsets $\phi^\ast {\mathcal{L}ip}_1(X)$ and $\psi^\ast {\mathcal{L}ip}_1(Y)$, taken over all possible parametrizations $\phi$ and $\psi$. Fig. \ref{fig:gromdist} illlustrates the concept.

\begin{theorem}[Gromov] A family $(X_n)$ of spaces with metric and measure is a L\'evy family if and only if $X_n$ converges in the distance $d_{conc}$ to a singleton $\{\ast\}$.
\end{theorem}

The closer a dataset $X$ is to a singleton $\{\ast\}$ in Gromov's distance, the higher its intrinsic dimensionality is and the more it resembles a ``black hole'' from the viewpoint of data analysis, because {\em all} the features simultenaously become less and less discriminaing. 
This reflects the fact that on a space of high intrinsic dimension the features are $\e$-contant on a set of measure $>1-2\alpha_X(\e)$, which is close to $1$ already for small values of $\e>0$. Consequently, the Hausdorff distance between ${\mathcal{L}ip}_1(X)$ and the set of functions on $\{\ast\}$ (that is, constant functions) is close to zero. 

\section{Main results}

\subsection{\label{ss:ax}Axiomatic approach to intrinsic dimension}

Let $\partial$ be a function assigning to every space $(X,d,\mu)$ with metric and measure either a non-negative real number or the symbol $+\infty$. We will say that $\partial$ is an {\em intrinsic dimension function} if it satisfies the following three axioms.

\subsubsection{\label{ax:conc}axiom of concentration} For a family $(X_n)$ of spaces with metric and measure,
$\partial(X_n)\uparrow\infty$ if and only if $(X_n)$ forms a L\'evy family.

This axiom formalizes a requirement that the intrinsic dimension is high if and only if a dataset suffers from the curse of dimensionality. 

\subsubsection{axiom of smooth dependence on datasets}  If\hfill $d_{conc}(X_n,X)\to 0$, then $\partial(X_n)\to \partial(X)$.

This axiom is necessary to assure that if a dataset $X$ is well-approximated by a non-linear manifold $M$, then the instrinsic dimension of $X$ is close to that of $M$. 

\subsubsection{axiom of normalization} $\partial(\s^n)=\Theta(n)$.\footnote{Recall that $f(n)=\Theta(g(n))$ if there exist constants $0<c<C$ and an $N$ with $c\abs{f(n)}\leq \abs{g(n)}\leq C\abs{f(n)}$ for all $n\geq N$. One says that the functions $f$ and $g$ asymptotically have the same order of magnitude.}

This axiom serves to properly calibrate the values of the intrinsic dimension.

\begin{remark}
Instead of spheres, one can use normalized hypercubes, Hamming cubes, Euclidean spaces with standard Gaussian distribution, etc. -- it can be proved that each of these families results in an equivalent definition.
\end{remark}
\vskip .2cm

The axioms immediately lead to a paradoxical conclusion. Since the Euclidean spheres $\s^n$ of radius one with the rotation-invariant probability measure form a L\'evy family \cite{MS,Gr}, they converge to a singleton $\{\ast\}$ with regard to Gromov's distance, and Axioms 1 and 2 (or 2$+$3) imply that
\[\partial(\{\ast\})=+\infty.\]

The converse is also true. Let $(X,d,\mu)$ be a space with metric and measure such that the support of $\mu$ is all of $X$.
\vskip .2cm

\begin{theorem} Let $\partial$ be an intrinsic dimension function. Then $\partial(X)=+\infty$ if and only if $X$ is a singleton: $X\simeq\{\ast\}$. 
\end{theorem}

\begin{proof}
If $\partial(X)=+\infty$, then the constant sequence $X_n=X$ is a L\'evy sequence, and so ${\mathrm{ObsDiam}}\,(X)=0$. This is only possible when $\mu$ is Dirac's point mass.
\end{proof}

Thus, the one and only infinite-dimensional object in a theory is a single point! 
This paradox seems to be unavoidable if one wants a notion of intrinsic dimension capable of detecting the curse of dimensionality, however it does not seem to lead to any problems or inconveniences.

Perhaps even more surprising is the fact that a dimension function satisfying the above requirements actually exists.

\subsection{Example: concentration dimension}

For an space with metric and measure $(X,d,\mu)$, define
\begin{equation}\label{eq:concdim}
\dim_{\alpha}(X)=\frac{1}{\left[2\int_0^1 \alpha_X(\e)~d\e\right]^2}.\end{equation}
We call $\dim_\alpha(X)$ the {\em concentration dimension} of $X$.

\begin{theorem}
The function $\dim_{\alpha}$ is an intrinsic dimension function. 
\end{theorem}

\begin{proof}
Axiom 1 follows at once from a standard result in Real Analysis (Lebesgue's Dominated Convergence Theorem). Axiom 2 involves a geometrical argument, to be published elsewhere. Axiom 3 is based on results obtained decades ago by Paul L\'evy \cite{Levy} (cf. also \cite{MS,L}). The inequality\footnote{Recall that $f(n)=\Omega(g(n))$ if for a constant $C>0$ and a natural $N$ one has $\abs{f(n)}\geq C\abs{g(n)}$ for all $n\geq N$. It is easy to see that the condition $f=\Theta(g)$ is equivalent to the conjunction of $f=O(g)$ and $f=\Omega(g)$.} $\dim_{\alpha}(\s^n)=\Omega(n)$ follows from a standard Gaussian upper bound on the concentration function of the sphere \cite{MS,L} 
\[\alpha_{\s^n}(\e)\leq C_1\exp(-C_2\e^2 n).\]
On the other hand, the value of concentration function $\alpha_{\s^n}(\e)$ is the relative $n$-volume of a spherical cap of height $1-\e$, and L\'evy's calculations show that in order for a spherical cap to keep a constant relative volume as $n\to\infty$, the height of such a cap should be on the order $\e=1-\Theta(1/\sqrt n)$. This suffices to obtain the other inequality: $\dim_{\alpha}(\s^n)=O(n)$.
\end{proof}

\begin{remark}
\label{r:bad}
One can replace $1$ with any fixed real number $L>0$ as the upper limit of integration in Eq. (\ref{eq:concdim}). It would be more natural to integrate to $+\infty$ and set 
\begin{equation}\label{eq:concdim2}
\dim_{\alpha}(X)=\frac{1}{\left[2\int_0^{\infty} \alpha(\e)~d\e\right]^2},\end{equation}
however Axiom 1 will no longer hold. Let $X=[1,+\infty)$ be a semi-infinite interval with the usual distance $d(x,y)=\abs{x-y}$ and probability density $p(x)=1/x^2$. Now one has
\[\alpha_X(\e)=\frac 1{2+\e},\]
so $\int_0^\infty\alpha_X(\e)\,d\e$ diverges to infinity.
The concentration dimension of such a space in the sense of Eq. (\ref{eq:concdim2}) is zero. One can
modify this example and obtain a L\'evy family of spaces with vanishing concentration dimension. Still, for all practical purposes it is more convenient to assume the definition in Eq. (\ref{eq:concdim2}) and restrict it to spaces with integrable concentration function (including, for instance, all spaces with bounded metric).
\end{remark}

Even if the concept of concentration dimension is introduced here for the first time, some known results can be reformulated in such a way as to underscore its theoretical relevance. Particular instances of the following theorem are well-known and often used, although in a different disguise (cf. \cite{MS}, p. 60), so we leave the proof out.

\begin{theorem}
\label{th:mm}
The median and the mean of a $1$-Lipschitz function $f$ on a space $(X,d,\mu)$ differ between themselves by at most $1/\sqrt{\dim_{\alpha}(X)}$ (in the sense of Eq. (\ref{eq:concdim2})). \hfill\QED
\end{theorem}

\begin{figure}[htp]
\centerline{\includegraphics[width=8.51cm]{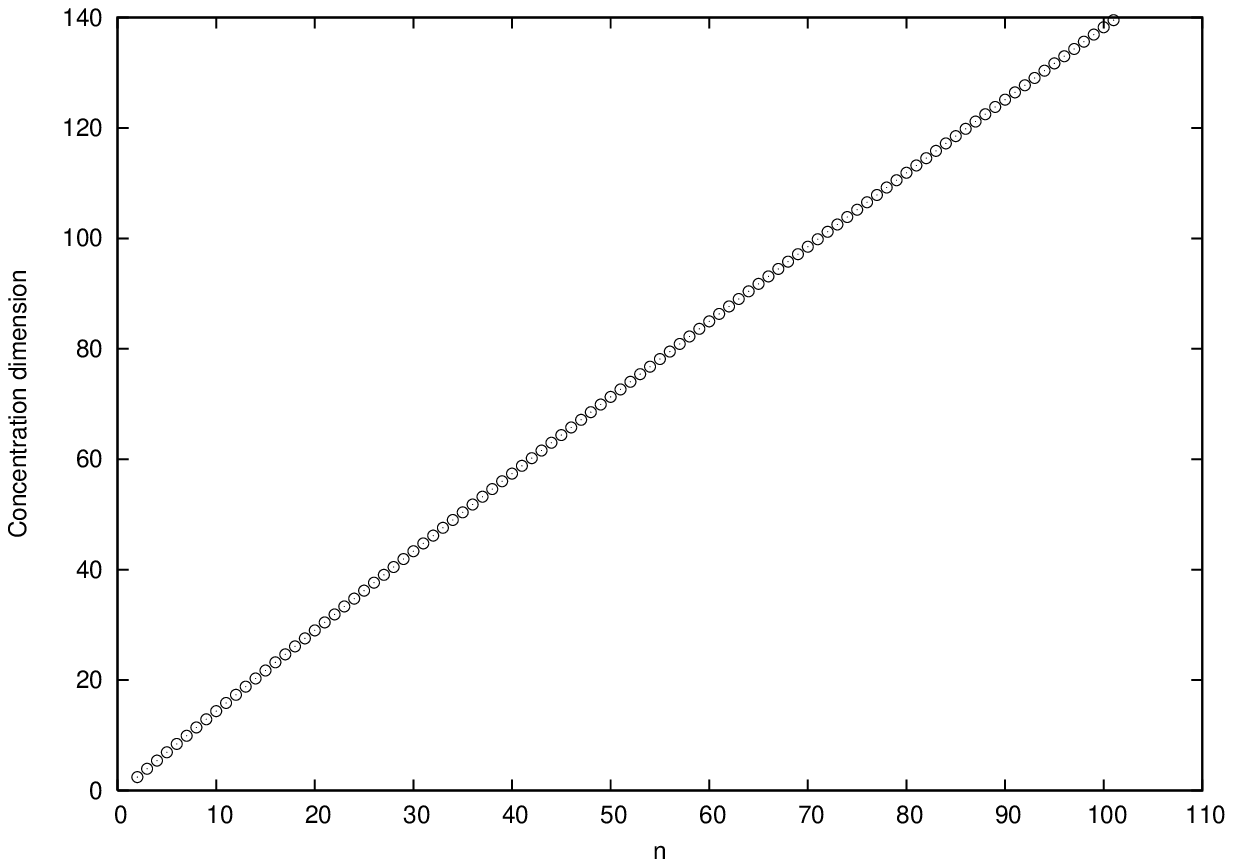}}
\caption{Concentration dimension of $n$-spheres for all $2\leq n\leq 101$.}
\label{fig:concD100}
\end{figure}

Euclidean spheres $\s^n$ of unit radius are among very few concrete families of geometric objects for which the exact values of $\dim_{\alpha}$ can be computed. (Cf. Fig. \ref{fig:concD100}.) 

\begin{example}
\label{ex:twospheres}
Let 
\[\s^{n-1}_i=\{x\in\R^{n+1}\colon x_1=i,~~x_2^2+x_3^2+\ldots+x_n^2=1\},\] where $i=0,1$, be two copies of the unit sphere $\s^{n-1}$ sitting inside $\R^{n+1}$ at a distance $1$ from and parallel to each other. Consider their union
\[X^n=\s^{n-1}_0\cup \s^{n-1}_1.\]
(Cf. Fig. \ref{fig:twospheres}.)

\begin{figure}[htp]
\centerline{\includegraphics[width=5cm]{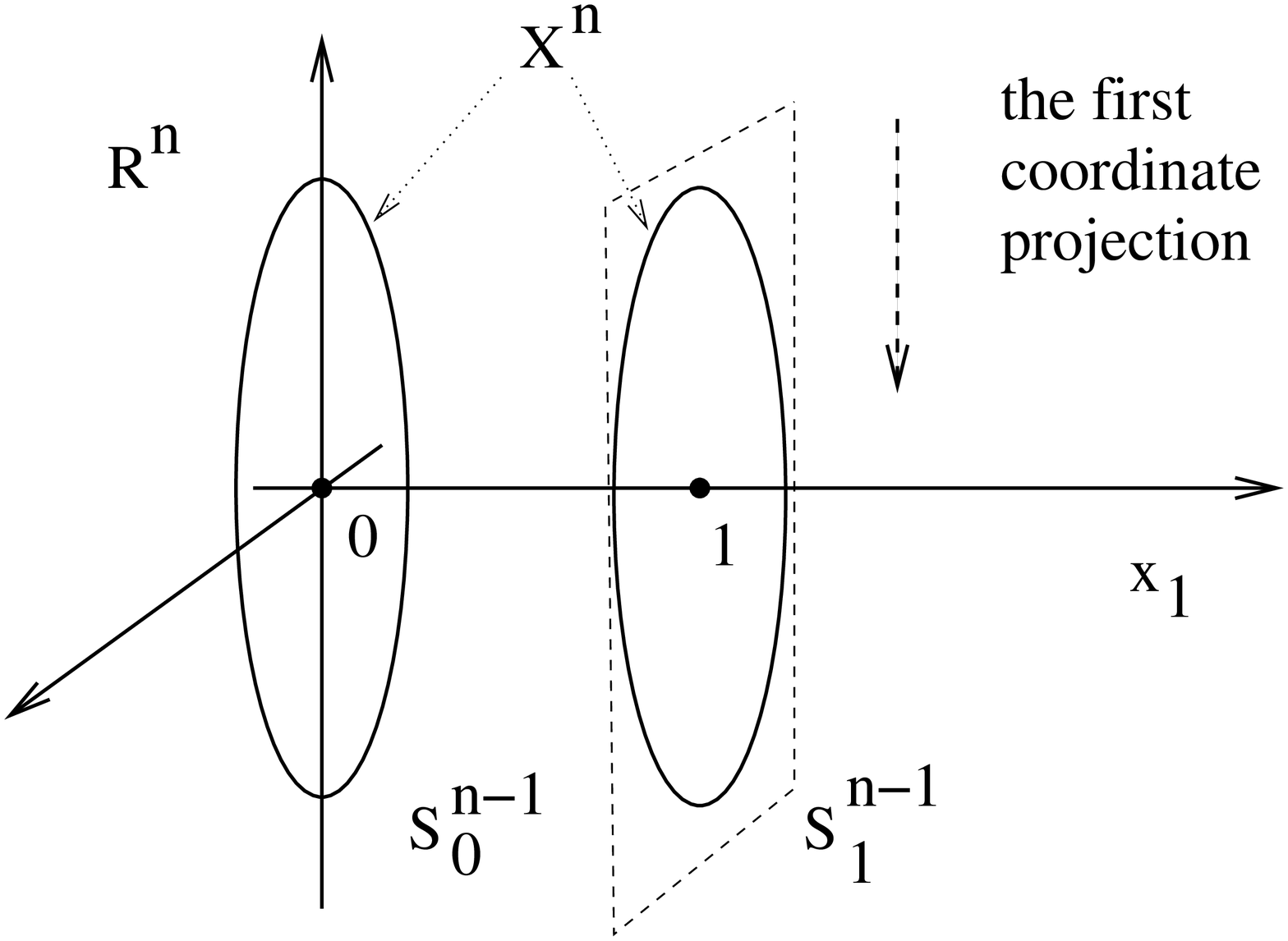}}
\caption{The space $X^n$ from Example \ref{ex:twospheres}.}
\label{fig:twospheres}
\end{figure}

Equip $X^n$ with the Euclidean distance coming from $\R^{n+1}$ and define a probability measure $\mu$ as follows: $\mu(A)=\mu^{(n-1)}(\s^{n-1}_0\cap A) + \mu^{(n-1)}(\s^{n-1}_1\cap A)$. (Here $\mu^{(n-1)}$ is the rotation-invariant measure on $\s^{n-1}$.)

Among all subsets $A$ of measure $\geq \frac 12$, those whose $\e$-neighbourhoods have the smallest measure are exactly the spheres $\s^{n-1}_i$, $i=0,1$, which form two well-separated clusters inside $X^n$. The concentration function of $X^n$ satisfies
\[\alpha_{X^n}(\e)=\left\{\begin{array}{ll}
\frac 12,&\mbox{ if }0\leq\e\leq 1,\\
0&\mbox{ otherwise,}\end{array}\right.\]
and $\dim_{conc}(X^n)=1$ for all $n$, another type of paradoxical behaviour!

This agrees with the fact that the sphere $\s^{n-1}$ of high dimension is close (in the Gromov distance) to a singleton, and therefore $X^n$ is close to the two-point space $\{0,1\}$. A low value of the concentration dimension indicates the existence of a well-separating feature: the first coordinate projection $X^n\to \{0,1\}$. 
\end{example}

\subsection{The intrinsic dimensionality of Ch\'avez et al.}

\begin{figure}[htp]
\centerline{\scalebox{0.22}[0.271]{\includegraphics{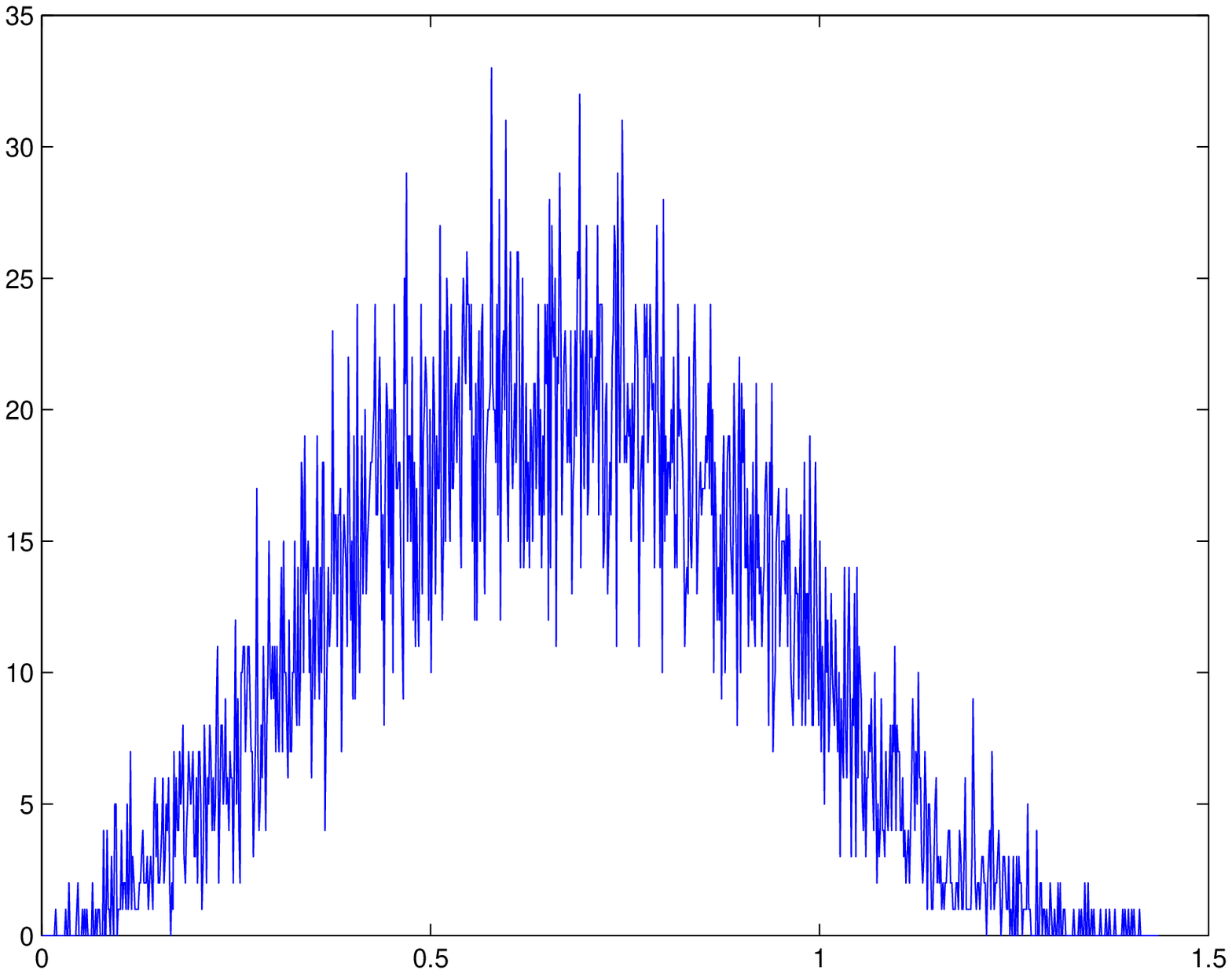}}
\hskip 1cm
\scalebox{0.22}[0.271]{\includegraphics{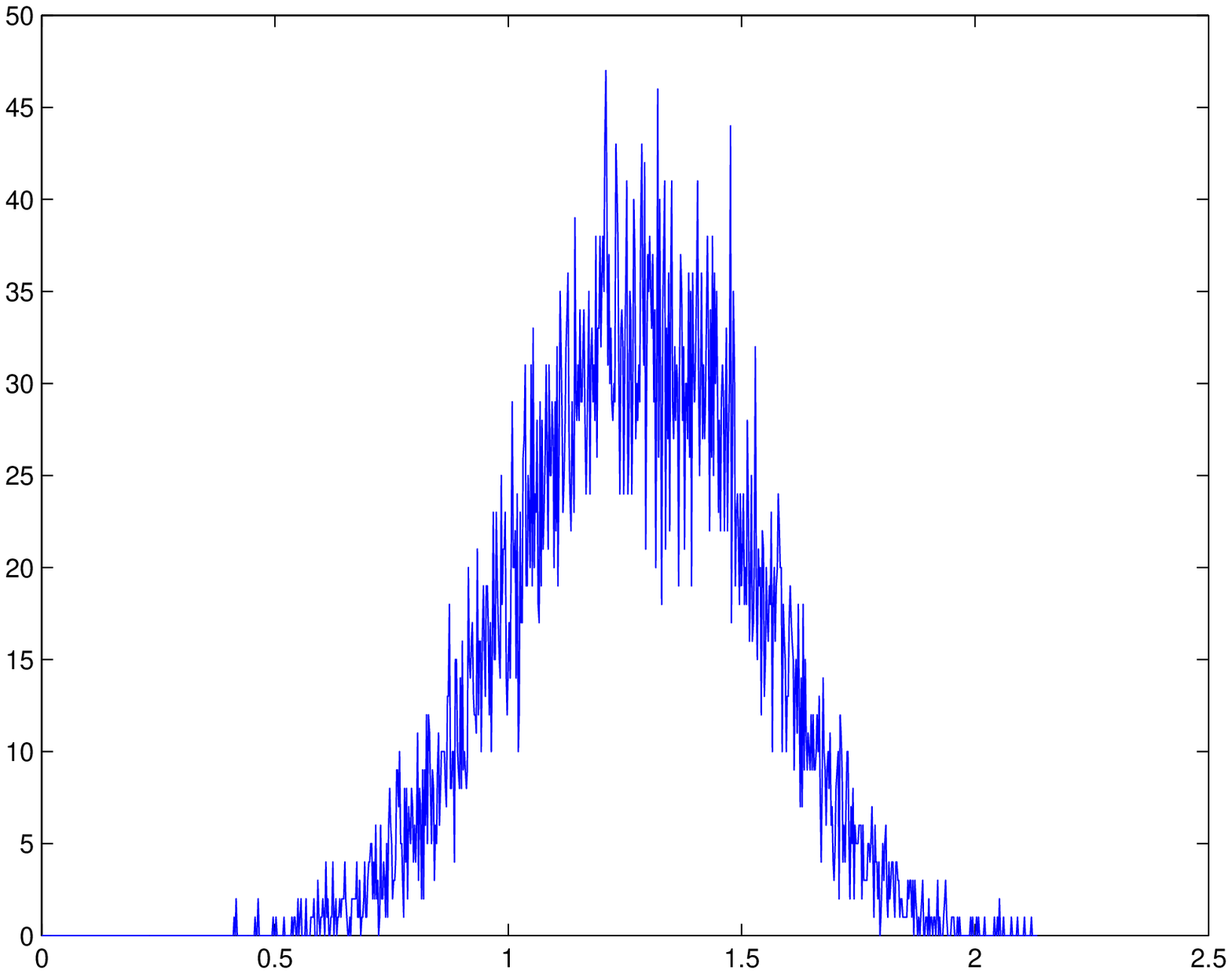}}}
\centerline{\scalebox{0.22}[0.271]{\includegraphics{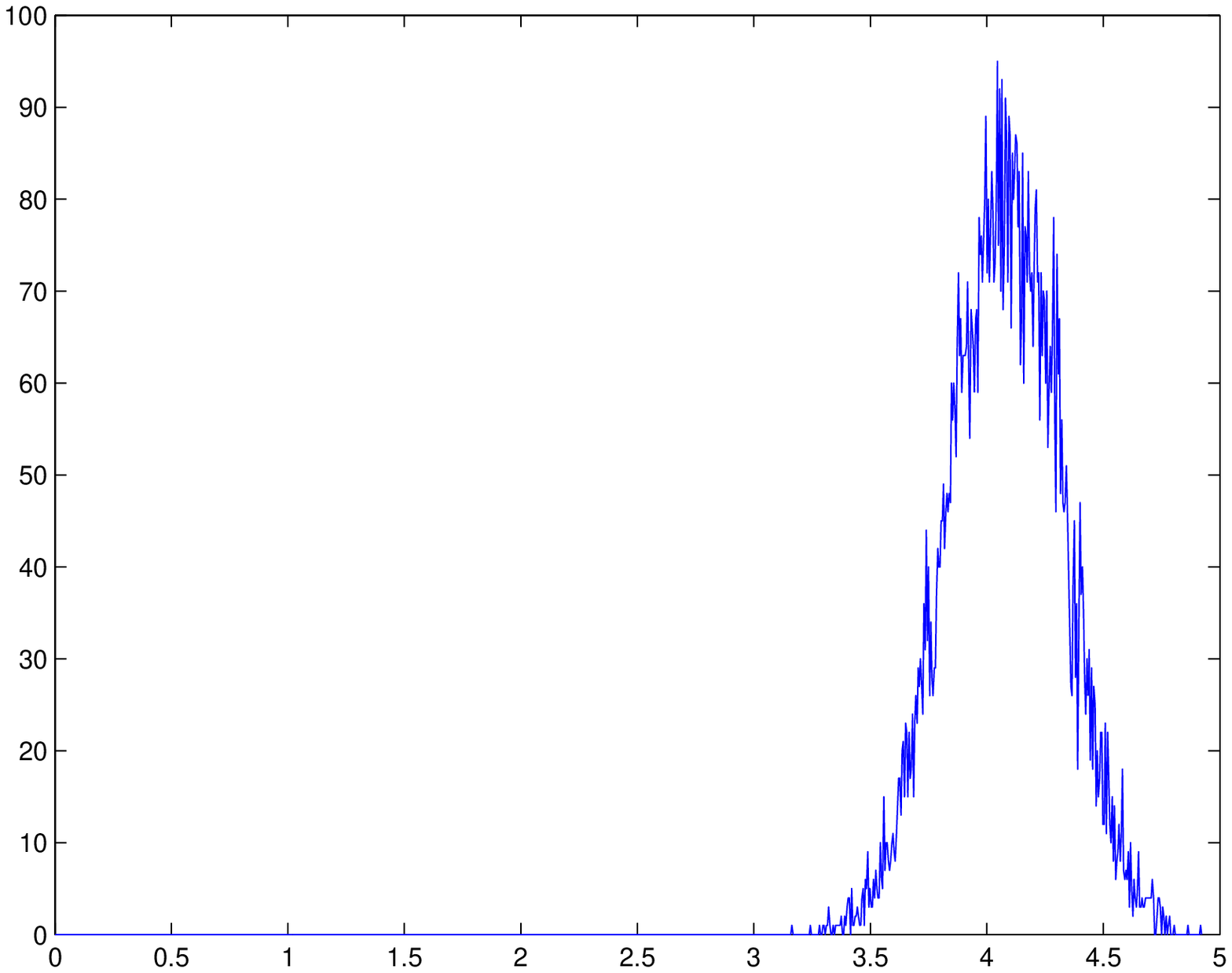}}
\hskip 1cm
\scalebox{0.22}[0.271]{\includegraphics{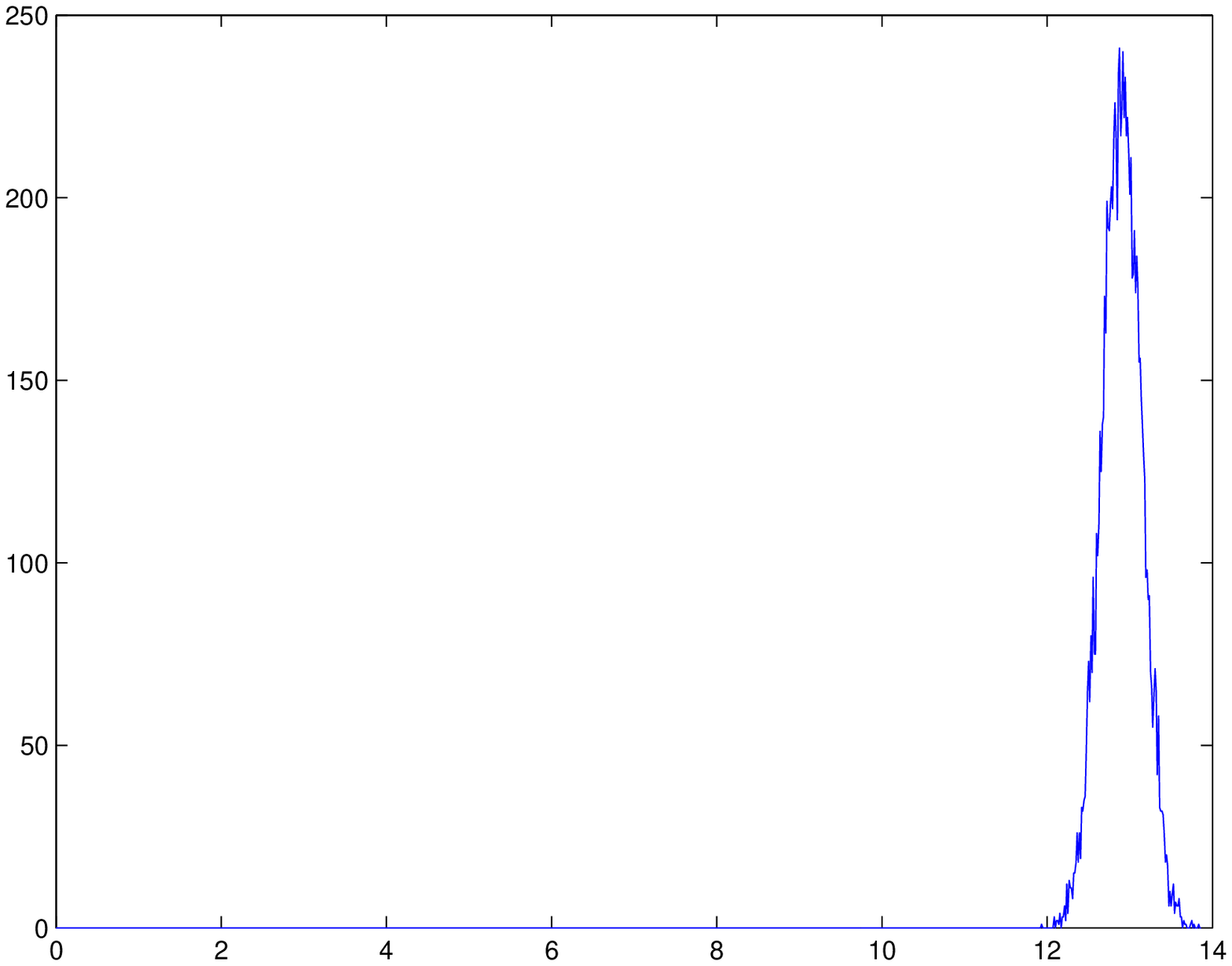}}}
\caption{Distribution of distances between
randomly chosen pairs of points in the unit hypercubecube $\I^n$, $n=3,10,100,1000$.
(Each histogram is based on a random sample of 10,000 pairs.)}
\label{fig:cdd}
\end{figure}

The following interesting version of intrinsic dimension was proposed by Ch\'avez {\em et al.} \cite{CNBYM} who called it simply {\em intrinsic dimensionality}. The concept explores a well-known property of high-dimensional spaces: the values of distances between points are sharply concentrated near one value (the characteristic size of $X$), cf. Fig. \ref{fig:cdd}.

Let $(X,d,\mu)$ be a space with metric and measure.
Denote by $m(d)$ the mean of the distance function $d\colon X\times X\to\R$ on the space $X\times X$ with the product measure. Assume $m(d)<\infty$. (This is not always the case: consider the space from Remark \ref{r:bad}.)
Let $\sigma(d)$ be the standard deviation of the same function. The intrinsic dimensionality of $X$ is defined as
\begin{equation}
\dim_{dist}(X)=\frac{m^2(d)}{2\sigma^2(d)}.
\label{eq:int}
\end{equation}

\begin{theorem}
The intrinsic dimensionality of Ch\'avez {\em et al.} satisfies:
\begin{itemize}
\item a weaker version of Axiom 1: if $(X_n,d_n,\mu_n)$ is a L\'evy family of spaces 
with bounded metrics, then $\dim_{dist}(X_n,\{\ast\})\to \infty$,
\item A weaker version of Axiom 2: if\hfill $d_{conc}(X_n,X)\to 0$ and $m(d_n)\to m(d)$, then $\dim_{dist}(X_n)\to \dim_{dist}(X)$,
\item Axiom 3.
\end{itemize}
\label{th:chavez}
\end{theorem}

\begin{proof} For the first property,
notice that
if $(X_n)$ is a L\'evy family, then so is $(X_n\times X_n)$, and the distance function $d_n$ concentrates near its median value, which can be replaced with the mean value by Theorem \ref{th:mm}.

The second property follows immediately from a similar property of the concentration dimension, while the proof of Axiom 3 uses symmetries of the sphere and is similar to the proof of Axiom 3 for the concentration dimension.
\end{proof}

\begin{remark}
For a singleton Eq. (\ref{eq:int}) returns $\frac 00$, and this value is genuinely undefined. Indeed, denote by $\e X_N$ a space with $N$ points at a distance of $\e$ from each other, equipped with the normalized counting measure. It is easy to see that 
\[\dim_{dist}(\e X_N)=\Theta(N)\to +\infty\mbox{ as }N\to\infty.\]
When $\e\to 0$, each of the spaces $\e X_N$ converges to a singleton in Gromov's distance, and so one cannot assign any particular value to the intrinsic dimension $\dim_{dist}(\{\ast\})$.
\end{remark}

This difference in behaviour is due to the fact that the intrinsic dimensionality is not an exact analogue of our concentration dimension, but rather of its normalized analogue 
${\dim_{conc}}(X)\times {\mathrm{CharSize}}(X)^2$. 

\begin{example}
\label{ex:XN}
The concentration function of the space $X_N=1\cdot X_N$ as above is easy to compute:
\[\alpha_{X_N}(\e)=\left\{\begin{array}{ll}\frac 12,&\mbox{ if }\e\leq 1, \\
0,&\mbox{ if }\e >1,
\end{array}\right.\]
and so $\dim_{conc}(X_N)=1$ for every $N$. At the same time, $\dim_{dist}(X_N)\to\infty$, even as ${\mathrm{CharSize}}\,(X_N)=\Theta(1)$.
\end{example}

One can argue that in Example \ref{ex:XN} the intrinsic dimensionality of Ch\'avez {\em et al.} gives away more useful information than the concentration dimension, because the spaces $X_N$ are often used to illustrate the curse of dimensionality in the context of similarity search as a toy example \cite{BGRS}. This case, which may or may not qualify as a genuine specimen of the ``curse of dimensionality'' (when finding nearest neighbours is easy, it just just outputting them all that is expensive), is indeed missed by our approach.

\begin{example} 
The intrinsic dimensionality of the spaces $X^n$ from Example \ref{ex:twospheres} (cf. Fig. \ref{fig:twospheres}) is uniformly bounded over all $n$. Indeed,
the mean distance between two random points $x,y\in X^n$ goes to $\sqrt 2$ as $n\to\infty$ provided $x,y$ are from the same sphere, and to $\sqrt 3$ otherwise. Since the two events are equiprobable, $m(d)\to(\sqrt 3 + \sqrt 2)/2$. Similarly, $\sigma^2(d)\to (\sqrt 3-\sqrt 2)^2/4$, and 
\[\dim_{dist}(X^n)\longrightarrow \left(\frac{\sqrt 3+\sqrt 2}{\sqrt 3 -\sqrt 2}\right)^2\approx 97.99\mbox{ as }n\to\infty.\]
\end{example}

\begin{table}[h]
\caption{Estimates of intrinsic dimensionality of spaces $X^n$ from Ex. \ref{ex:twospheres}}
\begin{center}
\begin{tabular}{|c||c|c|c|c|c|c|c|}
\hline
 $n$        &  $2$  & $3$&  $10$ &$30$ & $100$ & 1000 & 5000 \\ \hline
 $\dim_{dist}(X^n)$&$6.7$&$11.2$&$34.0$&$61.7$ &$83.5$&$96.3$ & $97.7$ \\ \hline
\end{tabular}
\label{tab:twosphere}
\end{center}
\end{table}

See Table \ref{tab:twosphere} for estimates of $\dim_{dist}(X^n)$ for selected values of $n$, based on the distance distribution of randomly sampled $3\cdot 10^5$ pairs (elements of $X^n\times X^n$). Keep in mind that the topological dimension of $X^n$ is $n-1$, while the concentration dimension is $1$.

\subsection{Some other approaches to instrinsic dimension} 
The approaches to intrinsic dimension listed below are all quite different both from our approach and from that of Ch\'avez {\em et al.}, in that they are set to emulate various versions of {\em topological} (i.e. essentially local) dimension.
All of them fail both our Axioms 1 and 2 and satisfy $\dim(X^n)=\Theta(n)$ for the two-sphere space $X^n$ from Example \ref{ex:twospheres}.

$\bullet$ {\em Correlation dimension,} which is a computationally efficient version of the box-counting dimension, see \cite{cv,tmgm}.

$\bullet$ {\em Packing dimension}, or rather its computable version as proposed and explored in \cite{Kegl}.

$\bullet$ {\em Distance exponent} \cite{ttf}, which is a version of the well-known Minkowski dimension.

$\bullet$ An algorithm for estimating the intrinsic dimension based on the Takens theorem from differential geometry \cite{pa}.

$\bullet$
A non-local approach to intrinsic dimension estimation based on entropy-theoretic results is proposed in \cite{ch}, however in case of  manifolds the algorithm will still return the topological dimension, so the same conclusions apply.

\section{Conclusions}

We have proposed a mathematical formalism for dealing with intrinsic dimension functions of datasets (as well as more general geometric objects) satisfying two requirements: a high intrinsic dimension is indicative of the curse of dimensionality, and closeness of two objects to each other implies the values of intrinsic dimension are also close. We formulate these conditions in a rigorous way, and demonstrate that a dimension function with such properties exists. We also discuss some of its paradoxical properties, such as, for instance, the infinite value of intrinsic dimension of a single point.

This dimension function, interesting as it may be, has two serious deficiencies. First, from the computational perspective it appears to be, generally speaking, untractable. Second, even 
if known, it need not be usable. A low value of dimension $\dim_\alpha$ indicates at an existence of a $1$-Lipschitz function $f$ on $X$ that is well dissipated (has high variance), and the corresponding ``geodesic flow'' gives a principal curve for $X$. However, it may happen that such an $f$ has very high complexity (examples are distance functions from large, complicated subsets of $X$). In applications, one is more interested in a situation where the features come from a specified class $\mathcal F$ of low-cost functions. (For example, in theory of indexing for similarity search, $\mathcal F$ may consist of distance functions to points.) Developing a corresponding concept of an intrinsic dimension function may solve both of the above problems, and here \cite{CNBYM} can serve as an important case study.

We also discuss the Gromov distance between spaces with metric and measure.
This distance {\em per se} is computationally even harder to estimate. However, notice that any intrinsic dimension function gives at least a qualitative estimate on the closeness of a dataset $X$ to the one-point space $\{\ast\}$. 
A similar estimate would be much more interesting and useful were a singleton replaced by a two point, or, better still, a $k$ point space (i.e., a singular principal manifold). This is an obvious next step to explore. Very likely, such estimates are already implicitely present in the great body of existing work on principal manifolds. 

\section*{Acknowledgments}
This work was supported in part by the NSERC discovery grant (2003-07) and by the University of Ottawa internal grants. Helpful comments from three anonymous referees are much appreciated.

\end{document}